\documentclass[letterpaper, 10 pt, conference, twocolumn]{ieeeconf}  

\IEEEoverridecommandlockouts                              

\usepackage{threeparttable}
\usepackage{cite}
\usepackage[ruled,vlined,linesnumbered]{algorithm2e}
\usepackage{algpseudocode}
\usepackage[utf8]{inputenc}
\usepackage[T1]{fontenc}
\usepackage{graphicx}
\usepackage{booktabs}
\usepackage{amssymb}
\usepackage{xcolor}
\usepackage{bm}
\let\labelindent\relax
\usepackage{enumitem}

\usepackage{cuted} 
\usepackage{caption}
\usepackage{amsmath}

\SetKwProg{Fn}{Function}{}{}

\newif\ifreview
\reviewfalse

\title{\LARGE \bf USS-Nav: Unified Spatio-Semantic Scene Graph for Lightweight UAV Zero-Shot Object Navigation}

\ifreview
    \author{Anonymous Author$^{1}$%
    \thanks{\phantom{$^{\dagger}$Corresponding authors: Fei Gao and Zhijun Meng}}%
    \thanks{\phantom{$^{1}$School of Aeronautic Science and Engineering, Beihang University, Beijing 100191, China.}}
    \thanks{The full text is omitted for double-blind review.} %
    \thanks{\phantom{$^{2}$State Key Laboratory of Industrial Control Technology, Zhejiang University, Hangzhou 310027, China.}}
    \thanks{\phantom{$^{3}$Differential Robotics, Hangzhou 311121, China.}}%
    \thanks{\phantom{E-mail: {\tt\ \{gaiwq, mengzhijun\}@buaa.edu.cn}}}%
    }
\else
    \author{Weiqi Gai$^{1, 3}$, Yuman Gao$^{2, 3}$, Yuan Zhou$^{2, 3}$, Yufan Xie$^{1}$, Zhiyang Liu$^{3}$, Yuze Wu$^{2, 3}$, Xin Zhou$^{2, 3}$ \\ Fei Gao$^{\dagger, 2, 3}$, and Zhijun Meng$^{\dagger, 1}$%
    \thanks{$^{\dagger}$Corresponding authors: Fei Gao and Zhijun Meng}%
    \thanks{$^{1}$School of Aeronautic Science and Engineering, Beihang University, Beijing 100191, China.}%
    \thanks{$^{2}$State Key Laboratory of Industrial Control Technology, Zhejiang University, Hangzhou 310027, China.}%
    \thanks{$^{3}$Differential Robotics, Hangzhou 311121, China.}%
    \thanks{E-mail: {\tt\ \{gaiwq, mengzhijun\}@buaa.edu.cn}}%
    \thanks{E-mail: {\tt fgaoaa@zju.edu.cn}}%
    }
\fi

\UseRawInputEncoding
\begin{document}

\maketitle

\begin{strip}
    \vspace{-2.0cm}
    \centering
    \includegraphics[width=\textwidth]{figs/top_figure.jpg}
    \captionof{figure}{Overview of the proposed framework deployed on a resource-constrained UAV for Zero-Shot Object Navigation in a real-world environment. The system incrementally constructs a Global Spatial Connectivity Graph visualized as a wireframe and concurrently instantiates semantic objects to form a unified spatio-semantic scene graph. This hierarchical representation supports dynamic region partitioning, enabling the Large Language Model for efficient, coarse-to-fine decision making.} 
    \label{fig:first-page-image}

\end{strip}

\thispagestyle{empty}
\pagestyle{empty}

\begin{abstract}

Zero-Shot Object Navigation in unknown environments poses significant challenges for Unmanned Aerial Vehicles (UAVs) due to the conflict between high-level semantic reasoning requirements and limited onboard computational resources. To address this, we present USS-Nav, a lightweight framework that incrementally constructs a Unified Spatio-Semantic scene graph and enables efficient Large Language Model (LLM)-augmented Zero-Shot Object Navigation in unknown environments. Specifically, we introduce an incremental Spatial Connectivity Graph generation method utilizing polyhedral expansion to capture global geometric topology, which is dynamically partitioned into semantic regions via graph clustering. Concurrently, open-vocabulary object semantics are instantiated and anchored to this topology to form a hierarchical environmental representation. Leveraging this hierarchical structure, we present a coarse-to-fine exploration strategy: LLM grounded in the scene graph's semantics to determine global target regions, while a local planner optimizes frontier coverage based on information gain. Experimental results demonstrate that our framework outperforms state-of-the-art methods in terms of computational efficiency and real-time update frequency (15 Hz) on a resource-constrained platform. Furthermore, ablation studies confirm the effectiveness of our framework, showing substantial improvements in Success weighted by Path Length (SPL). The source code will be made publicly available to foster further research.

\end{abstract}
\section{Introduction}

Driven by the inherent agility and superior mobility of multirotor Unmanned Aerial
Vehicles (UAVs), their demand in real-world applications is rapidly increasing \cite{sun2024survey}. Incorporating semantic perception and understanding enables UAVs to transcend simple point-to-point obstacle avoidance, unlocking the ability to autonomously explore and reason about previously unseen spaces. This capability is pivotal for Zero-Shot Object Navigation, empowering a UAV to execute natural-language commands such as "find a sofa in the living room" or "locate a suitable place for sleeping" without any prior knowledge of the scene layout.


Prior works are commonly demonstrated in simulation \cite{deitke2020robothor} or on ground robots \cite{gu2024conceptgraphs,zhang2025apexnav} with plentiful compute and power. In contrast, UAVs operate under strict size, weight, and power (SWaP) constraints, making such heavy pipelines difficult to deploy and run in real time. To make Zero-Shot Object Navigation practical on such constrained aerial platforms, we identify two critical challenges:
\textbf{a)} Lack of a unified spatial-semantic representation, and \textbf{b)} Prohibitive computational burdens imposed by limited onboard resources. Existing methods struggle to efficiently bridge these gaps. While traditional occupancy-based strategies \cite{zhou2021fuel} excel at geometric coverage, they lack the semantic grounding necessary for high-level interaction. Conversely, Scene Graphs \cite{hughes2022hydra, gu2024conceptgraphs, yin2024sg, chen2025irs} provide hierarchical abstractions suitable for LLMs but often decouple from intrinsic geometry and lack spatial geometric memory. This necessitates heavy region partitioning algorithms and additional dense maps for global navigation, resulting in high computational overhead and low efficiency in long-horizon tasks~\cite{chen2025irs}. Although Semantic Value Maps \cite{zhang2025apexnav, zhou2025beliefmapnav} and Vision-Language Navigation (VLN) methods \cite{du2025vl} offer task-specific solutions, they typically suffer from limited reasoning capabilities in multi-task scenarios or lack persistent memory for long-horizon exploration. Crucially, the reliance on computationally intensive models in these approaches often mandates offloading the entire computational workload to external servers, rendering them impractical for fully autonomous, resource-constrained UAVs.

To address these issues, we propose a framework for the progressive synthesis of hierarchical scene graphs in unexplored environments to facilitate semantic-aware exploration. Specifically, we first generate Spatial Connectivity Graphs via polyhedral expansion, directly embedding global geometric navigability. Based on this topology, we apply topological graph clustering to achieve lightweight regional segmentation, resulting in a topological hierarchical structure. Subsequently, by leveraging lightweight visual foundation models to extract salient objects, we construct an object-centric scene graph online (up to 15 Hz on Jetson Orin NX), integrating these semantic elements with the geometric and regional layers. To further optimize computational overhead, we introduce a coarse coverage grid for memory-efficient tracking of known/unknown regions, which allows the system to maintain high-resolution dense maps only locally. Ultimately, this hierarchical scene graph structure enables coarse-to-fine decision-making, leading to highly efficient semantic exploration. 

We evaluate the proposed system through extensive experiments in high-fidelity simulations and complex, large-scale real-world scenarios, demonstrating its superior performance in addressing the aforementioned challenges. The contributions of this paper are summarized as follows:
\begin{itemize}[leftmargin=*]
\item Incremental spatial connectivity graph generation via polyhedral expansion, and subsequent graph clustering for spatial region partitioning.
\item An online 3D object-centric scene graph construction method, featured with hierarchical semantic information, accurate geometry and light computational cost.
\item An LLM-augmented exploration framework integrating scene graphs and semantic reasoning to support downstream object navigation.
\item Development and deployment of a lightweight Zero-Shot Object Navigation system on a resource-constrained UAV, achieving real-time performance validated through experiments in both high-fidelity simulations and real-world scenarios. The source code and simulation environment will be made publicly available to foster further~research.
\end{itemize}

\section{RELATED WORK}

\subsection{Object Navigation}
Object navigation frameworks are broadly categorized into end-to-end and modular paradigms. Data-driven end-to-end methods~\cite{du2021vtnet,pal2021learning} map raw observations directly to actions but often require extensive environment-specific training, limiting their generalization. Recently, the integration of LLMs and Vision-Language Models (VLMs) has revolutionized this field, leveraging common-sense reasoning to achieve remarkable Zero-Shot and Open-Vocabulary performance~\cite{yin2024sg,yokoyama2024vlfm,zhang2025apexnav,han2025muvla,zhou2025beliefmapnav}. However, the significant computational burden of these large models, combined with complex environmental representations, hinders their direct deployment on resource-constrained UAVs. Adhering to a modular philosophy, we propose a lightweight framework that decouples perception and reasoning, integrating efficient foundation models to enable practical onboard Zero-Shot Object Navigation.

\subsection{Scene Graph Construction}
Graph-based environmental representations have gained prominence for their ability to structure complex spatial data~\cite{liu2025aligning}. Early approaches~\cite{zhu2022nice} relied on SLAM to generate closed-vocabulary 3D semantic maps. With the advent of foundation models~\cite{ren2024grounded,liu20233d}, focus has shifted to open-vocabulary scene graphs. State-of-the-art methods like ConceptGraphs~\cite{gu2024conceptgraphs} and SG-Nav~\cite{yin2024sg} utilize graph structures to organize high-level semantics and inter-object relationships. To enhance spatial reasoning, some works have introduced hierarchical abstractions, such as room-level segmentation via grid expansion~\cite{hughes2022hydra} or online point cloud segmentation~\cite{chen2025irs}. 

Despite these advancements, extracting essential spatial relationships remains computationally prohibitive for aerial platforms, often requiring offline processing or powerful desktop GPUs. Addressing this bottleneck, we propose a lightweight, multi-level scene graph grounded in the efficient \textit{ikd-tree} structure. We concurrently extract a spatial skeleton topology via polyhedral expansion for rapid region clustering, enabling online construction on UAVs while utilizing asynchronous LLM processing for environmental narration.

\subsection{Zero-Shot Object Navigation Strategy}
Research on Zero-Shot Object Navigation for UAVs remains nascent. Traditional UAV exploration strategies predominantly focus on geometric coverage using frontier-based~\cite{yamauchi1997frontier} or global-optimal algorithms~\cite{zhou2021fuel}, which ignore semantic context and are inefficient for specific object search tasks. Conversely, semantic-aware methods for ground robots often rely on heavy Vision-Language-Action models or employ LLMs to score frontiers without explicit spatial memory~\cite{zhang2024trihelper,zhang2025apexnav}, making them unsuitable for high-frequency aerial control.

Our approach bridges the gap between geometric efficiency and semantic reasoning. We propose a hierarchical strategy: at the global level, an LLM selects exploration regions based on the scene graph's semantics; locally, a Traveling Salesperson Problem (TSP) solver optimizes frontier traversal. This design decouples high-level decision-making from low-level path planning, ensuring robustness and computational feasibility for real-time UAV deployment.

\section{SYSTEM OVERVIEW}

The proposed framework leverages rolling local occupancy grid maps and RGB-D observations as inputs. As illustrated in Fig.~x, the system first generates a Global Spatial Connectivity Graph to provide structural global guidance (Sec.~IV-A). Based on the generated topological nodes, a graph-based incremental region partitioning algorithm partitions the space to derive region layer information (Sec.~IV-B). Concurrently, a lightweight algorithm performs spatially-consistent semantic object instantiation to complete the Unified Spatio-Semantic Scene Graph seamlessly integrating grid, topology, region, and object layers while initializing semantic descriptions (Sec.~IV-C). Subsequently, the semantic-aware hierarchical exploration is initiated. This module combines a Global Coverage Mask (GCM) with local occupancy data for GCM-aided frontier extraction (Sec.~V-A) and employs an LLM for coarse-to-fine decision-making, grounded in the scene graph's semantics (Sec.~V-B). The mission concludes when the LLM determines that the semantic information fully satisfies the given instruction requirements.

\begin{figure*}[t] 
    \centering
    \includegraphics[width=\textwidth]{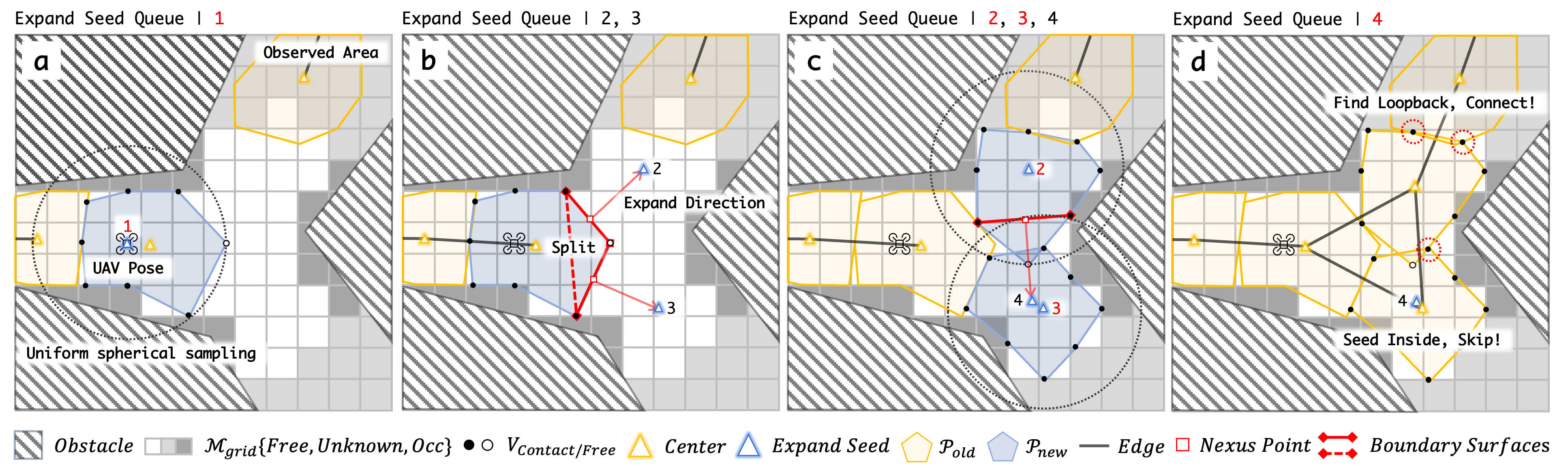}
    \caption{Illustration of the incremental Spatial Connectivity Graph generation pipeline. Sub-figures (a) through (d) sequentially demonstrate the algorithmic workflow, evolving from uniform spherical sampling and polyhedral boundary expansion to the final topological update. The process constructs the Global Spatial Connectivity Graph directly from the current local occupancy grid map.} 
    \label{fig:skeleton_gen}
\end{figure*}

\section{Unified Spatio-Semantic Scene Graph}

\begin{table}[t]    \centering
    \caption{Nomenclature and Symbol Definitions}
    \label{tab:symbols}
    \renewcommand{\arraystretch}{1.3}
    \begin{tabular}{ll}
        \toprule
        \textbf{Symbol} & \textbf{Explanation} \\
        \midrule
        $\mathcal{M}_{grid}$ & Local occupancy grid map \\
        $\mathcal{M}_{SCG}$ & Global Spatial Connectivity Graph \\
        \hspace{1em} $E$ & Edges of the topological structure \\
        \hspace{1em} $\mathcal{R}$ & Segmented Region \\
        \hspace{1em} $\mathcal{P}$ & Polyhedron node within $\mathcal{M}_{SCG}$ \\
        \hspace{2em} $\mathcal{S}$ & Surfaces (Faces) of Polyhedron $\mathcal{P}$ \\
        \hspace{2em} $\mathcal{V}$ & Vertices of Polyhedron $\mathcal{P}$ \\
        \hspace{2em} $\mathcal{N}$ & Geometric Center (Centroid) of Polyhedron $\mathcal{P}$ \\
        \bottomrule
    \end{tabular}
\end{table}

\subsection{Global Spatial Connectivity Graph Generation}

To enable high-quality global navigation within unexplored environments under strict onboard constraints, we necessitate a globally sparse spatial representation. Drawing inspiration from \cite{chen2022fast}, where topological graphs derived from spatial skeletons effectively characterize spatial connectivity, we identify a critical limitation in its direct application: the reliance on global batch processing of point clouds renders it unsuitable for incremental updates on grid-based local maps. To bridge this gap, we propose a method that incrementally constructs a Global Spatial Connectivity Graph. Our approach utilizes polyhedra to approximate free space, autonomously determining growth directions via Boundary Surfaces to ensure geometric consistency with a minimal memory footprint.

For clarity, the key nomenclature and symbols used throughout this section are defined in Table \ref{tab:symbols}.

\begin{algorithm}[t]
\caption{Spatial Connectivity Graph Updation}
\label{alg:skeleton}
\DontPrintSemicolon
\small

\KwIn{$\mathcal{M}_{grid}$, current position $p_{cur}$}
\KwOut{$\mathcal{M}_{SCG}$}

\SetKwProg{Fn}{Function}{:}{}
\SetKwFunction{FUpdate}{UpdateTopology}
\SetKwFunction{FExpand}{ExpandNewPolyhedron}
\SetKwFunction{FFind}{FindNewExpandSeed}

\SetKwFunction{FLinkParent}{ConnectToParent}
\SetKwFunction{FLinkColl}{ConnectCollidingNbrs}
\SetKwFunction{FLinkVis}{ConnectVisibleNbrs}

    $p_{seed} \leftarrow p_{cur}$ \;
    $\mathcal{P}_{init} \leftarrow$ \FExpand{$p_{seed}$} \;
    $\mathcal{M}_{SCG}.\text{Add}(\mathcal{P}_{init})$ \;
    
    \While{$\bm{\mathcal{S}}_{boundary} \neq \emptyset$}{
        $\mathcal{S} \leftarrow \bm{\mathcal{S}}_{boundary}.\text{Pop}()$ \; 
        $p_{seed} \leftarrow$ \FFind{$\mathcal{S}$} \;
        $\mathcal{P}_{new} \leftarrow$ \FExpand{$p_{seed}$} \;
        $\bm{E}_{new} \leftarrow$ \FUpdate{$\mathcal{P}_{new}, r_{vis}$} \;
        $\mathcal{M}_{SCG}.\text{Append}(\mathcal{P}_{new}, \bm{E}_{new})$ \;
    }

\BlankLine
\hrule
\BlankLine

    \Fn{\FUpdate{$\mathcal{P}_{new}, r_{vis}$}}{
        $\bm{E}_{parent} \leftarrow$ \FLinkParent{$\mathcal{P}_{new}$} \;
        $\bm{E}_{col} \leftarrow$ \FLinkColl{$\mathcal{P}_{new}$} \;
        $\bm{E}_{vis} \leftarrow$ \FLinkVis{$\mathcal{P}_{new}, r_{vis}$} \;
        \KwRet $\bm{E}_{parent} \cup \bm{E}_{col} \cup \bm{E}_{vis}$ \;
    }
    
\end{algorithm}

To achieve rapid incremental updates and comprehensive spatial coverage, the generation process is formulated in Algorithm \ref{alg:skeleton}, which is divided into two primary phases: 1) Iterative Polyhedral Expansion (Lines 1-9), and 2) Topological Connection and Loop Closure (Lines 10-14). Fig.~\ref{fig:skeleton_gen} (a)-(c) illustrates the expansion logic, while Fig.~\ref{fig:skeleton_gen} (d) depicts the topological connection and update.

\subsubsection*{1) Iterative Polyhedral Expansion}
The algorithm accepts the local occupancy grid map $\mathcal{M}_{grid}$ and the UAV's current position $p_{cur}$ as input. The process initiates by designating $p_{cur}$ as the initial seed $p_{seed}$. To guarantee robust initialization, a primary polyhedron is instantiated at this origin to bootstrap the expansion process (Algorithm \ref{alg:skeleton}, Lines 1-3). Given the seed point $p_{seed}$, the algorithm executes the current iteration of Global Spatial Connectivity Graph generation (Algorithm \ref{alg:skeleton}, Lines 4-9) as follows:

\begin{itemize}[leftmargin=*]
    \item \textbf{Vertex Sampling:} Within \texttt{ExpandNewPolyhedron}, we perform uniform sampling on a unit sphere centered at $p_{seed}$. By performing ray-casting in $\mathcal{M}_{grid}$, we identify vertices $\mathcal{V}$ categorized as either \textit{Contact} (obstacles) or \textit{Free} (open space), as shown in Fig.~\ref{fig:skeleton_gen}(a).
    
    \item \textbf{Polyhedron Generation:} The QuickHull algorithm is employed to generate the set of surfaces $\mathcal{S}$ and construct the polyhedron $\mathcal{P}_{new}$.
    
    \item \textbf{Growth Direction Analysis:} To populate the expansion queue $\bm{\mathcal{S}}_{boundary}$, we analyze the surfaces of $\mathcal{P}_{new}$. Surfaces containing heterogeneous vertices indicate potential openings. These are clustered to identify \textbf{Boundary Surfaces}, shown as red edges in Fig.~\ref{fig:skeleton_gen}(b). A cluster is split if the number of constituent surfaces exceeds a threshold or if normal vectors deviate significantly.
    
    \item \textbf{Next Seed Determination:} To propagate expansion, the function \texttt{FindNewExpandSeed} calculates a \textbf{Nexus Point} by projecting the geometric center of a frontier cluster onto its representative plane. The next seed is generated by extending outwards from this point along the surface normal. The candidate seed undergoes a validity check (ensuring safety margins and non-enclosure) before the next iteration.
\end{itemize}

This process is iterated until boundary surface queue is exhausted, ensuring complete coverage of local free space.

\subsubsection*{2) Topological Update and Loop Closure}
Upon generating a new polyhedron, the function \texttt{UpdateTopology} (Algorithm \ref{alg:skeleton}, Lines 10-14) establishes the connectivity $E$ of the Spatial Connectivity Graph by aggregating three distinct types of connections:

\begin{itemize}[leftmargin=*]
    \item \textbf{Kinematic Connection (\texttt{ConnectToParent}):} A mandatory link is established with the parent polyhedron from which the current node was directly expanded, thereby ensuring kinematic reachability.
    \item \textbf{Structural Overlap (\texttt{ConnectCollidingNbrs}):} Connections are formed with existing nodes that spatially intersect with the current polyhedron, thereby maintaining strict geometric consistency.
    \item \textbf{Local Reachability (\texttt{ConnectVisibleNbrs}):} To facilitate loop closure, we connect to nodes within a visibility radius $r_{vis}$ that are either directly visible or connected via a collision-free path.
\end{itemize}

\subsection{Graph-Based Incremental Region Partitioning}

\begin{figure}[t]
    \centering
    \includegraphics[width=1.0\linewidth]{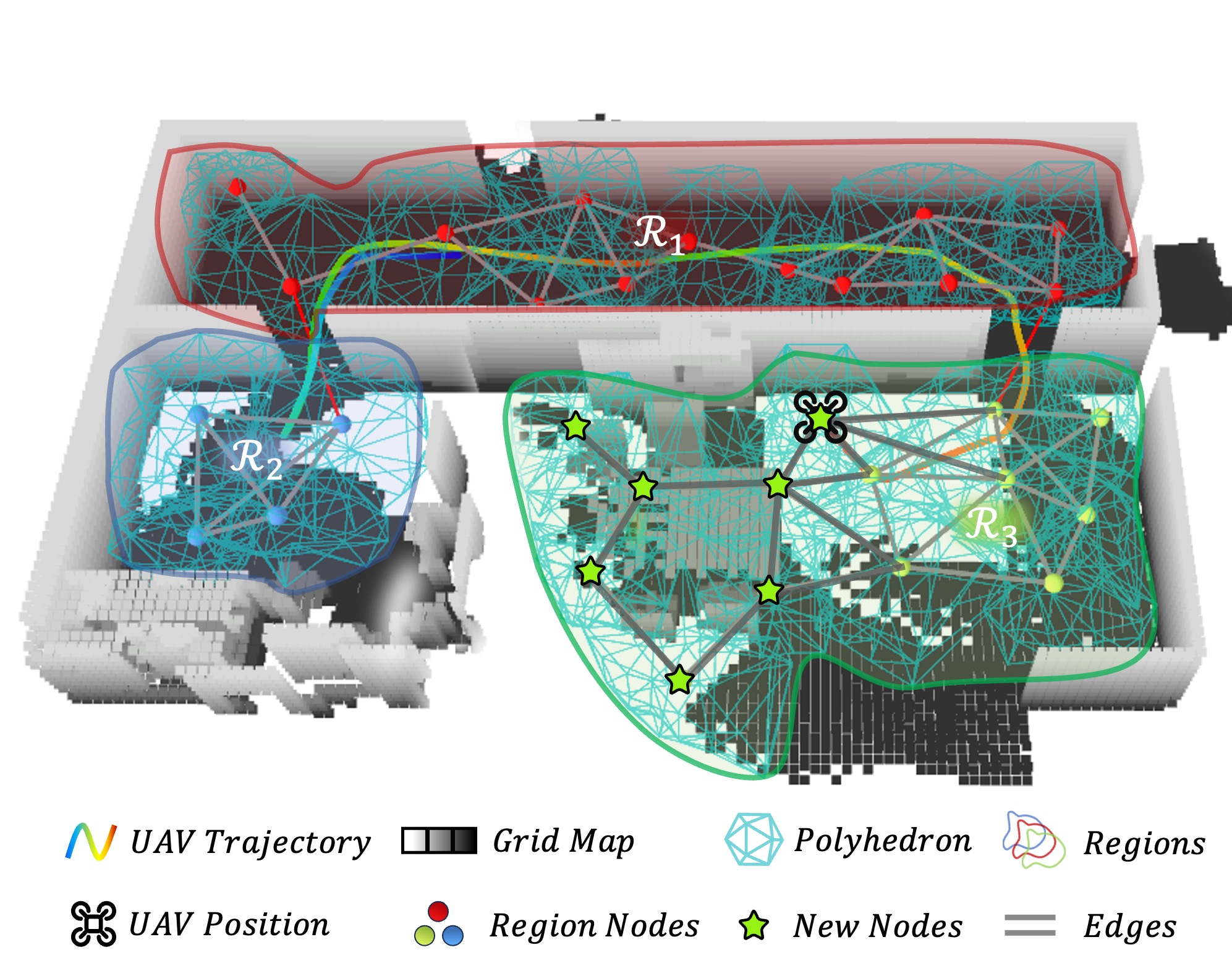 }
    \caption{Demonstration of the proposed spatial representation. This visualization showcases the effective polyhedral expansion for covering free space and the resulting region partitioning. It illustrates the dynamic process where newly observed nodes are seamlessly fused into established regions, ensuring topological consistency.}
    \label{fig:region_part}
\end{figure}

To facilitate hierarchical environmental understanding, we reformulate region segmentation as a topological clustering problem. Since the proposed $\mathcal{M}_{SCG}$ encapsulates both spatial geometry and inter-node connectivity, we can effectively partition the space by analyzing the community structure of the graph.

\subsubsection{Problem Formulation: Community Detection}
We construct an undirected, weighted graph where the vertices correspond to the polyhedral centroids $\mathcal{N}$ derived in Sec.~IV-A, and the edges correspond to the topological connections~$E$. To capture the distinct spatial relationships, we assign edge weights based on connection types: standard spatial connections are assigned a weight of 1, while "forced" connections (e.g., kinematic links without clear geometric adjacency) are assigned a small constant $\epsilon$ to discourage clustering across weak boundaries.

The segmentation task is thus framed as a Community Detection problem, where the objective is to partition the set of centroids $\mathcal{N}$ into disjoint regions $\bm{\mathcal{R}} = \{\mathcal{R}_1, \dots, \mathcal{R}_k\}$. The goal is to maximize the \textit{Modularity} of the partition, ensuring that nodes within the same region are densely connected while connections between different regions are sparse. To solve this optimization problem efficiently, we employ the Leiden algorithm~\cite{traag2019louvain}, which guarantees well-connected communities and avoids the resolution limit problem common in traditional methods.

\subsubsection{Region Incremental Update Algorithm}
Naive global re-clustering is computationally expensive and leads to inconsistent Region IDs over time. To address this, we propose an incremental algorithm that operates locally on newly generated data, as illustrated in Fig.~\ref{fig:region_part}. The process comprises two key stages:

\begin{itemize}[leftmargin=*]
\item \textbf{Local Subgraph Extraction:} To minimize computational overhead while preserving sufficient topological context, we extract a local candidate graph. As illustrated in Fig.~\ref{fig:region_part}, this subgraph is constructed by aggregating the newly generated polyhedra $\mathcal{P}_{new}$ (green stars) with both the immediate region $\mathcal{R}_3$ and its \textbf{entire adjacent neighbor $\mathcal{R}_1$}. This inclusion of full neighboring regions is essential for accurate community detection. Crucially, this approach remains strictly local spatially disjoint areas (e.g., $\mathcal{R}_2$) are excluded thereby reducing calculation costs and maintaining segmentation stability across the global map.

\item \textbf{Local Partitioning and Consolidation:} The Leiden algorithm \cite{traag2019louvain} is applied exclusively to this candidate graph. In the example, optimization is confined to the union of $\mathcal{P}_{new}$, $\mathcal{R}_3$, and the $\mathcal{R}_1$ boundary, allowing new nodes to seamlessly merge into $\mathcal{R}_3$ without perturbing spatially disjoint areas (e.g., $\mathcal{R}_2$ remains unaffected). Finally, a consolidation step merges adjacent regions with high connectivity density to mitigate over-segmentation, yielding the refined set $\mathcal{R}_{new}$.

\end{itemize}

\subsection{Spatially-Consistent Semantic Object Instantiation}

While the Spatial Connectivity Graph and region partitioning (Sec. IV-A $\&$ B) establish the fundamental environmental skeleton, effective semantic navigation necessitates fine-grained object-level perception. In this subsection, we detail the construction of the object layer, which robustly anchors open-vocabulary semantics to the spatial topology via a lightweight, spatially-consistent instantiation pipeline.

\begin{figure}[t]
    \centering
    \includegraphics[width=1.0\linewidth]{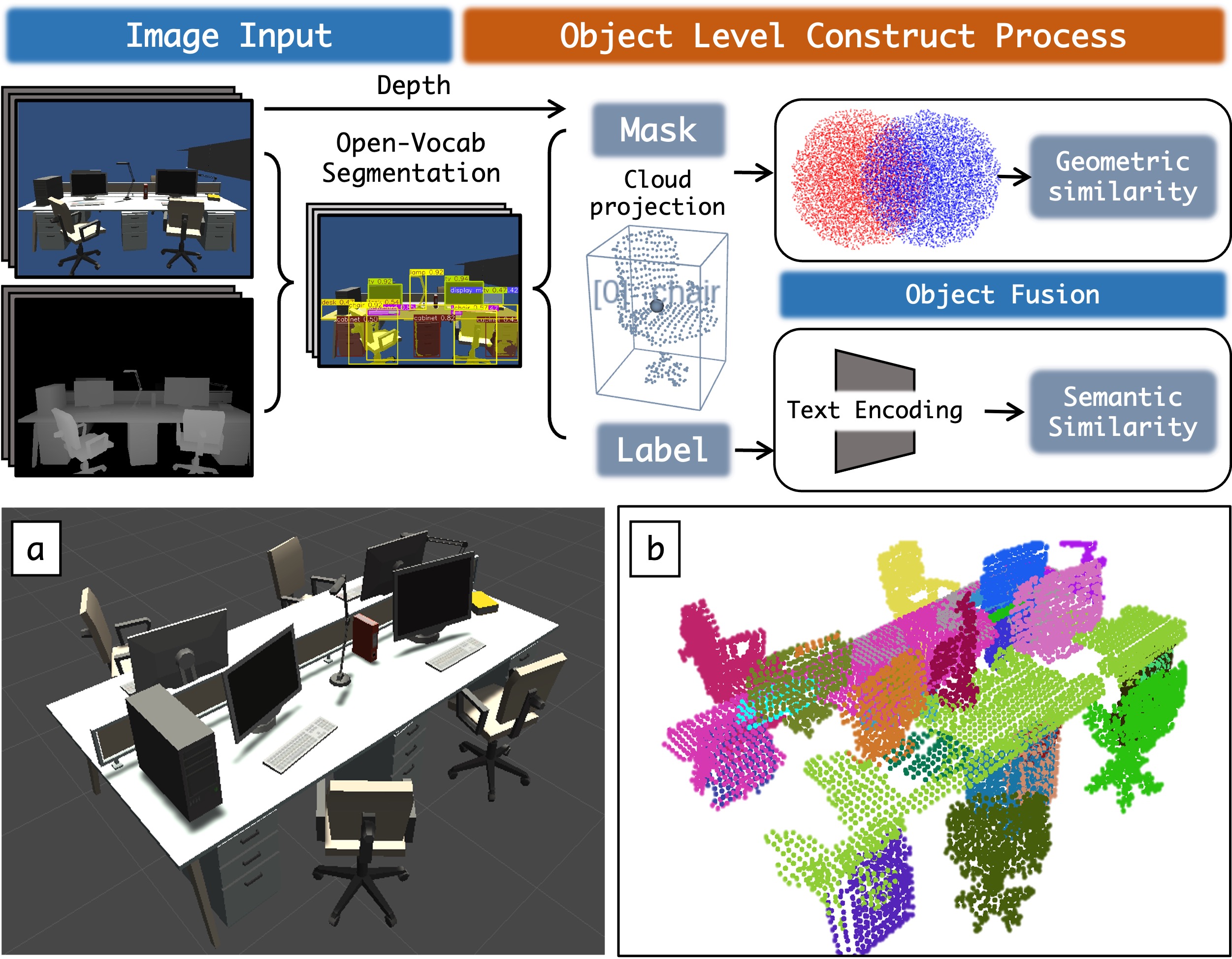}
    \caption{Framework for associating object semantics with spatial information in the scene graph, leveraging advanced open-vocabulary segmentation models combined with spatial and semantic similarity measures. \textbf{(a)} Ground truth in the simulation environment. \textbf{(b)} Object processing results.}
    \label{fig:obj_process}
\end{figure}

As illustrated in Fig.~\ref{fig:obj_process}, the pipeline processes a synchronized data sequence $S=\{(I_k, D_k, T_k)\}_{k=1}^{N}$, where $I_{k}$ represents the RGB image, $D_{k}$ denotes the aligned depth map, and $T_{k} \in SE(3)$ indicates the UAV's 6-DoF pose. The instantiation procedure unfolds as follows:

(1) An object-centric segmentation model YOLO-E \cite{wang2025yoloe} processes $I_k$ to yield instance masks $M_{i} \in \mathbb{R}^{H\times W}$ and class labels $l_{i}$. Simultaneously, Mobile-CLIP\cite{vasu2024mobileclip} encodes $l_i$ into a high-dimensional semantic vector $\mathbf{v}_{i} \in \mathbb{R}^{512}$.

(2) The inputs $(I_k, D_k, T_k)$ are processed in parallel. The masked pixels $(u,v)$ of each detected object are back-projected via intrinsics $\pi^{-1}$ using the depth $D_k$ and transformed to the world frame via the pose $T_k$. This generates the object point cloud, denoted here as $\mathcal{C}_i$.

(3) New observations are associated with the global scene graph. We query the ikd-tree for candidate objects within a 2m radius of the centroid of $\mathcal{C}_i$. Association relies on two distinct metrics.

First, the \textbf{Geometric Similarity} $\Omega_{geo}$ quantifies the spatial consistency between the new cloud $\mathcal{C}_i$ and an existing cloud~$\mathcal{C}_j$:

\begin{equation}
    \Omega(\mathcal{C}_{i}, \mathcal{C}_{j}) = \frac{1}{|\mathcal{C}_{i}|} \sum_{\mathbf{p} \in \mathcal{C}_{i}} \mathbb{I}\left( \min_{\mathbf{q} \in \mathcal{C}_{j}} \| \mathbf{p} - \mathbf{q} \| \leq \tau \right)
\end{equation}

\noindent
where $\mathbb{I}(\cdot)$ is the indicator function, and $\tau$ represents the Euclidean distance threshold for considering two points as a valid geometric match.

Second, the \textbf{Semantic Similarity} $\Omega_{sem}$  is computed as the cosine similarity between their feature vectors:

\begin{equation}
    \Omega_{sem} = \frac{ \mathbf{v}_{i}^\top \mathbf{v}_{j} }{ \| \mathbf{v}_{i} \| \| \mathbf{v}_{j} \| }
\end{equation}

\noindent
Fusion adheres to a prioritized hierarchy:

\begin{itemize} [leftmargin=*]
\item \textbf{Semantic Match:} Requires high semantic alignment ($\Omega_{sem}(\mathbf{v}_{i}, \mathbf{v}_{j})>0.75$) supported by moderate geometric overlap ($\Omega_{geo}(\mathcal{C}_{i}, \mathcal{C}_{j}) > 0.1$).
\item \textbf{Geometric Match:} In the absence of a strong semantic match, we accept candidates with significant bidirectional geometric overlaps ($\Omega_{geo}(\mathcal{C}_{i}, \mathcal{C}_{j}) > 0.5$ and $\Omega_{geo}(\mathcal{C}_{j}, \mathcal{C}_{i}) > 0.5$).
\end{itemize}

Matched objects trigger an incremental update, while unmatched observations instantiate new nodes. Crucially, each object is spatially anchored to the topological node corresponding to the UAV's current location.

\section{SEMANTIC-AWARE HIERARCHICAL EXPLORATION AND PLANNING}

\subsection{Global Coverage Mask-Aided Frontier Extraction}

To achieve resource-efficient large-scale exploration while providing geometric grounding for LLM-based decision-making, we propose a frontier generation framework.
Notably, this method relies solely on a local rolling map $\mathcal{M}_{grid}$ with a minimal memory footprint, augmented by a \textbf{Global Coverage Mask}, denoted as $\mathcal{M}_{GCM}$, to maintain global exploration consistency.

As illustrated in Fig.~\ref{fig:gcm_image}, the $\mathcal{M}_{GCM}$ is explicitly designed to compensate for the loss of historical global information inherent in local rolling maps. Each cell in $\mathcal{M}_{GCM}$ represents a fixed 3D spatial volume and tracks the number of unknown voxels within that region. The value of a GCM cell at index $\mathbf{k}$ is formally defined as:

\begin{equation}
V_{GCM}(\mathbf{k}) = \sum_{v \in \mathcal{V}(\mathbf{k}) \cap \mathcal{M}_{grid}} \mathbb{I}(S(v) = \text{unknown})
\end{equation}

\noindent where $\mathcal{V}(\mathbf{k})$ denotes the set of voxels contained within the cell's spatial domain, and $\mathbb{I}(\cdot)$ is the indicator function. A cell is marked as \textit{visited} if its unknown voxel count falls below a predefined threshold, thereby excluding it from subsequent frontier computations.

Leveraging an incremental update strategy~\cite{zhou2021fuel}, we extract frontiers by identifying boundaries between free and unknown spaces solely within the current sensor field of view. We then utilize $\mathcal{M}_{GCM}$ to filter out candidates located in \textit{visited} regions. For the remaining valid frontiers, we compute optimal viewpoints and information gain. Crucially, each generated frontier is anchored to the spatial skeleton node corresponding to the UAV's current position. This association effectively assigns frontiers to specific topological regions, serving as the basis for the region-level decision-making detailed in Sec.~V-B. Moreover, it enhances the computational efficiency of path search by allowing the planner to operate on the sparse skeleton structure.

\subsection{LLM-Driven Coarse-To-Fine Navigation}

\begin{figure}[t]
    \centering
    \includegraphics[width=1.0\linewidth]{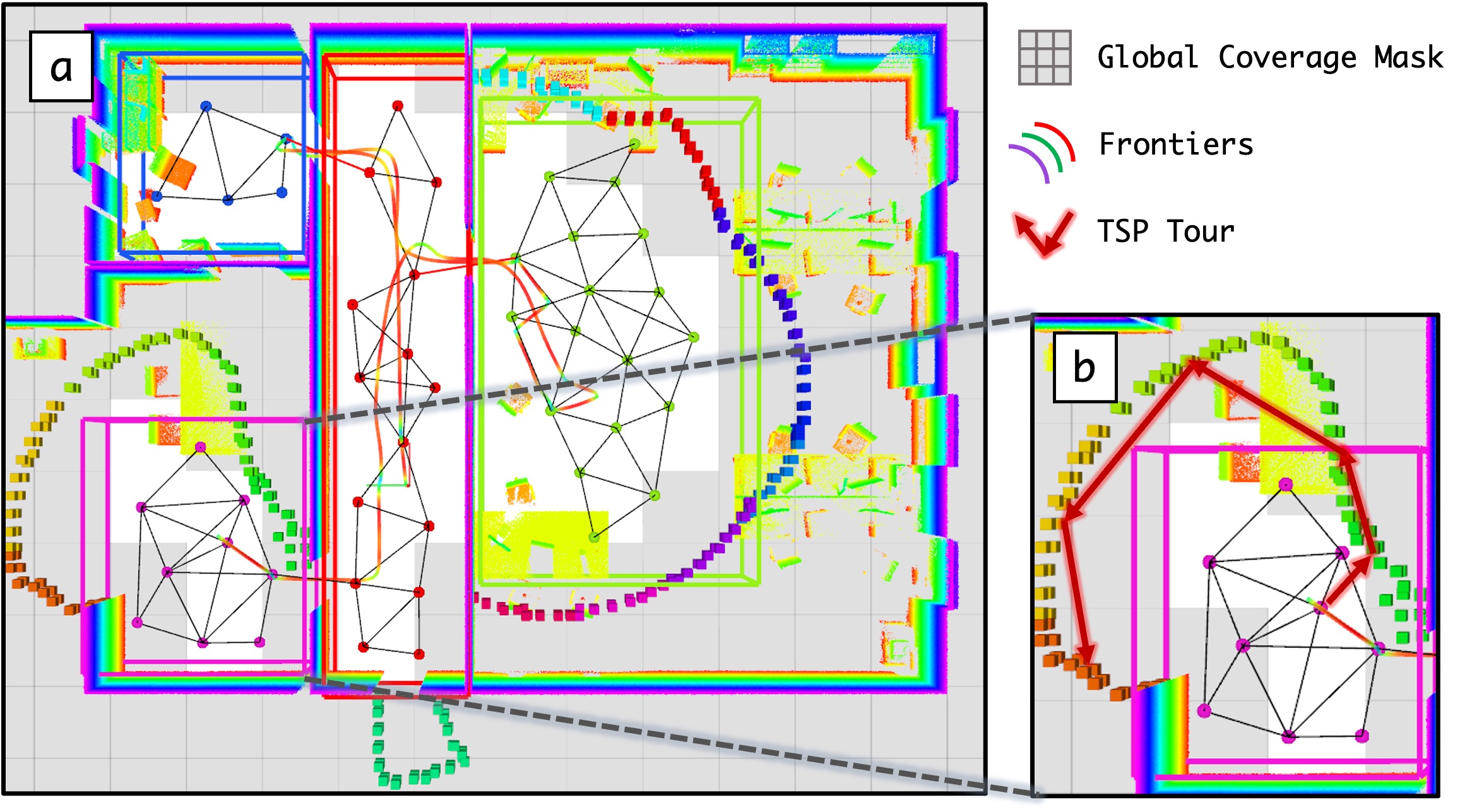}
    \caption{Frontier generation and regional association. 
\textbf{(a)} Frontier generation utilizing the Global Coverage Mask. White regions indicate visited status, suppressing subsequent frontier generation in these areas. 
\textbf{(b)} After the LLM-based coarse-level global reasoning, a TSP solver is employed within the target region for fine-level local planning.}
    \label{fig:gcm_image}
\end{figure}

The proposed methodology adopts a hierarchical coarse-to-fine paradigm to integrate LLM-based reasoning with traditional algorithmic robustness. We designate frontiers as atomic primitives and utilize the scene graph as a semantic bridge, enabling the system to transition from high-level semantic reasoning to low-level precise execution.

To support this hierarchy, we first implement an adaptive semantic extraction mechanism. Prior to decision-making, the system monitors changes in spatial structure and object-level semantics. Upon detecting variations, fine-grained object attributes \texttt{Label}, \texttt{Position}, and \texttt{Size} are serialized into JSON format. The LLM aggregates these cues to update the region semantic information, grounding the subsequent coarse-level reasoning.

The navigation process then unfolds in two distinct stages:

\textbf{1) Coarse-Level Global Reasoning:} We replace conventional global selection with an LLM-driven decision process. Key spatial data is serialized into a schema containing \texttt{Current Area}, \texttt{Visit History}, and \texttt{Regional Description}. Leveraging reasoning with region adjacency priors, the LLM synthesizes topology and semantics to identify the optimal \textbf{target region}.

\textbf{2) Fine-Level Local Planning:} Once the coarse target region is determined, the system shifts to fine-grained execution. We formulate the intra-region navigation as a TSP. A cost matrix is generated based on the information gain and path costs of all atomic frontiers within the selected region, yielding an optimal visitation sequence, \textbf{as illustrated in Fig.~\ref{fig:gcm_image}(b).}

In scenarios where semantic context is insufficient for coarse-level reasoning, the system seamlessly reverts to a global utility maximization strategy, prioritizing coverage expansion until mission requirements are met.

\begin{figure*}[t] 
    \centering
    \includegraphics[width=\textwidth]{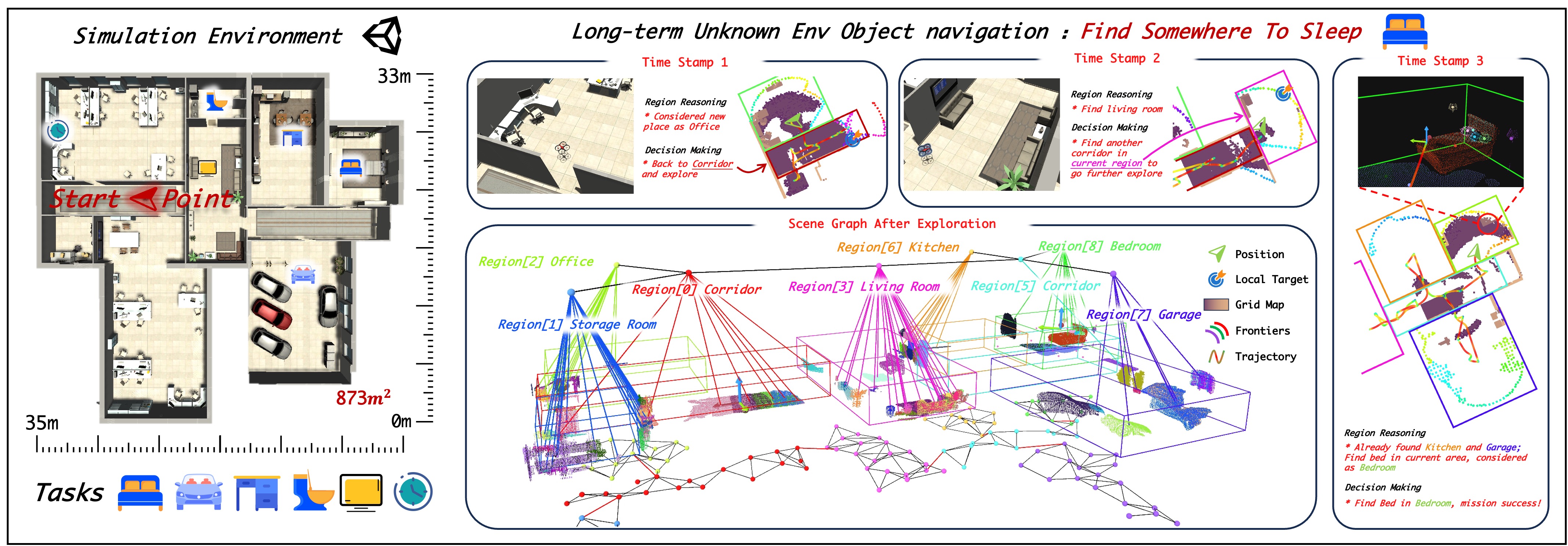}
    \caption{\textbf{Visualization of the simulation setup and task execution workflow.} The \textbf{left panel} illustrates the high-fidelity environment layout with annotated start and target locations for diverse navigation tasks. The \textbf{right panel} details the execution of a long-horizon task, where the scene graph is constructed incrementally. \textbf{Time Stamps 1-2} demonstrate the UAV's capability for online regional reasoning and hierarchical decision-making in varying spatial contexts. \textbf{Time Stamp 3} depicts the ego-centric semantic point cloud and successful target identification upon mission completion.}
    \label{fig:sim_exp}
\end{figure*}

\section{Experiments}

Research on Object Navigation in unknown environments for UAVs remains in a nascent stage, with the majority of existing studies relying on ground-based platforms. This reliance poses significant challenges for transferring algorithms to aerial robots due to the inherent differences in dynamics and viewing perspectives. Consequently, this section first details the task execution within a high-fidelity simulation environment, followed by a comparative evaluation against state-of-the-art methods. Finally, we present ablation studies and real-world experiments to validate the efficacy and robustness of our proposed framework.

\subsection{Simulation Experiment}

\begin{figure}[t]
    \centering
    \includegraphics[width=1.0\linewidth]{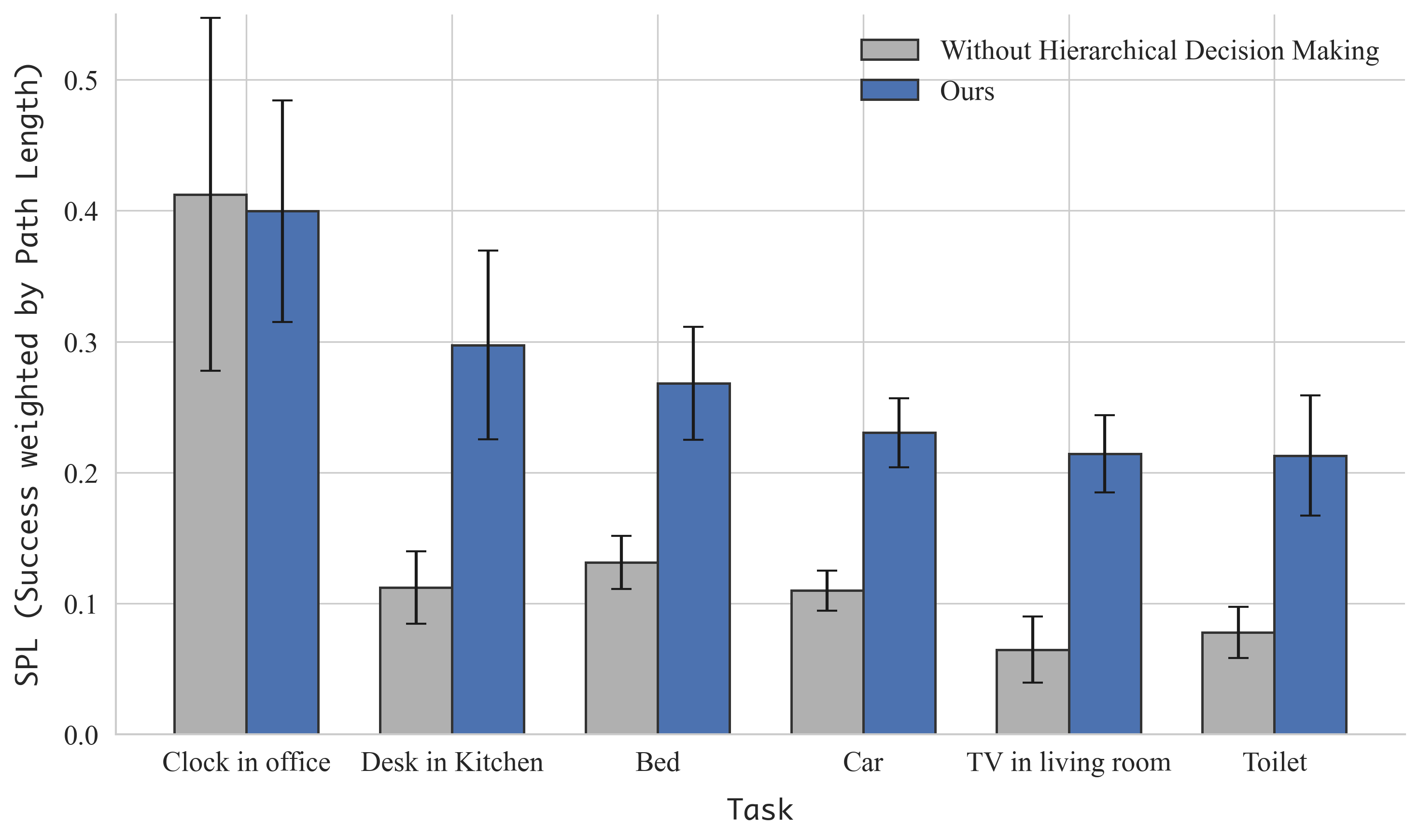}
    \caption{Ablation study results on the effectiveness of the hierarchical decision-making module.}
    \label{fig:ablation_polt}
\end{figure}

\subsubsection{Simulation Platform}
Simulation experiments are conducted on a workstation equipped with an Intel Core i7-10875H CPU and an NVIDIA RTX 2060 GPU. To address the limitations of common indoor simulators (e.g., Habitat), which often exhibit geometric artifacts in aerial views and lack accurate multirotor dynamics, we developed a high-fidelity 3D simulation environment based on the Marsim lightweight simulator and the Unity engine. This platform supports realistic sensor simulation tailored to our real-world hardware setup, ensuring reliable performance validation. An overview of this environment and a qualitative demonstration of the task execution workflow are presented in Fig.~\ref{fig:sim_exp}.

\subsubsection{Implementation Details}

We employ a fixed-memory rolling local occupancy grid map with dimensions of $8 \times 8 \times 4$ m and a resolution of 0.1 m. For polyhedral expansion, candidate vertices are sampled using a Fibonacci sphere with 50 points and a maximum radius of 1.4 m. The maximum seed extension distance is set to 2.0 m. In the region partitioning module, the resolution parameter for the Leiden algorithm is set to 0.02. 

Regarding the perception models, YOLO-E is accelerated using the TensorRT inference framework (FP16 mode), while Mobile-CLIP is executed directly using the BLT model. The LLM integration utilizes the official API of Qwen-3, which serves as the only module running on an external server; all other computations are performed locally. To ensure effective global guidance, we sparsely inject prior knowledge of spatial adjacency, such as connection between corridor and office, encoded as a room adjacency list.

\subsubsection{Hierarchical Decision Making Ablation Study}

To validate the effectiveness of the proposed Zero-Shot Object Navigation framework, we compare our full system against a baseline without hierarchical coarse-to-fine decision-making, where the UAV relies solely on global frontier utility. We selected diverse tasks ranging from single-room searches to multi-region traversal (locations marked in Fig.~\ref{fig:sim_exp}), conducting 5 trials per task. Performance is evaluated using \textbf{Success weighted by Path Length (SPL)}.

As shown in Fig.~\ref{fig:ablation_polt}, our method demonstrates superior robustness in complex scenarios. While the baseline performs similarly in simple, topologically trivial tasks (e.g., ``Clock in office''), its efficiency degrades sharply in long-horizon missions due to the lack of global semantic guidance. Conversely, our method utilizes the scene graph for high-level reasoning, significantly reducing ineffective exploration. Consequently, we achieve an average SPL improvement of \textbf{146.61\%} compared to the baseline in complex tasks, proving the necessity of the proposed hierarchical pipeline for efficient large-scale navigation.

\subsubsection{Comparison with State-of-the-art Methods}

\begin{table}[t]
    \centering
    \caption{Comparison with State-of-the-Art Methods.}
    \label{tab:sota_compare}
    \resizebox{\linewidth}{!}{%
    \begin{threeparttable}
        \begin{tabular}{lcccccc}
            \toprule
            \textbf{Method} & \textbf{Unk.} & \textbf{Reas.} & \textbf{GPS} & \textbf{Comp. Platform} & \textbf{Freq.} & \textbf{Robot} \\
            \midrule
            Hydra~\cite{hughes2022hydra} & $\times$ & $\times$ & $\times$ & Xavier NX (Edge) & 5 Hz & Sim \\
            ConceptGraph~\cite{gu2024conceptgraphs} & $\times$ & \checkmark & $\times$ & Offline & - & UGV \\
            IRS~\cite{chen2025irs} & $\times$ & \checkmark & $\times$ & RTX 4090 (PC) & Offline & UGV \\
            BeliefMapNav~\cite{zhou2025beliefmapnav} & \checkmark & $\times$ & \checkmark & RTX 4090 (PC) & $<1$ Hz & Sim \\
            ApexNav~\cite{zhang2025apexnav} & \checkmark & $\times$ & \checkmark & RTX 4060 (PC) & 3 Hz & UGV \\
            SG-Nav~\cite{yin2024sg} & \checkmark & $\times$ & \checkmark & - & $<1$ Hz & Sim \\
            \midrule
            \textbf{Ours} & \checkmark & \checkmark & \checkmark & \textbf{Orin NX (Edge)} & \textbf{15 Hz} & \textbf{UAV} \\
            \bottomrule
        \end{tabular}
        \begin{tablenotes}
            \footnotesize
            \item \textbf{Unk.}: Unknown Environment; \textbf{Reas.}: Reasoning Capability; 
            \item \textbf{GPS}: Global Path Searching; \textbf{Freq.}: Map Update Frequency.
        \end{tablenotes}
    \end{threeparttable}
    }
\end{table}

We perform a comprehensive comparison against state-of-the-art environmental representation frameworks designed for modular object navigation. The evaluation focuses on the intrinsic properties of the representation, including: its support for autonomous exploration in unknown environments, its capacity to facilitate high-level semantic reasoning, its enablement of global path planning without redundant auxiliary maps, and the associated computational overhead (platform type and update frequency).

As presented in Table~\ref{tab:sota_compare}, our proposed Unified Spatio-Semantic Scene Graph uniquely integrates these capabilities into a cohesive structure. Unlike approaches that rely on heavy desktop-grade GPUs (e.g., RTX 4090) or offline processing to maintain dense representations, our framework is sufficiently lightweight to run on an embedded edge platform (Orin NX) at a high frequency of 15 Hz, making it suitable for agile UAV applications.

\subsection{Real-World Deployment}

\subsubsection{Real-world Platform}
We validated our system on a custom-designed quadrotor platform. The UAV is equipped with an NVIDIA Jetson Orin NX (16GB RAM) onboard computer, which handles the online construction of the scene graph and motion planning modules. For state estimation and dense mapping, we utilize a Livox Mid-360 LiDAR running the Fast-LIO2 algorithm. A RealSense D455 camera provides RGB-Depth aligned image sequences for semantic scene graph generation.

\subsubsection{Unknown Environment Object Navigation}
We conducted Unknown Environment Object Navigation experiments in a large-scale, complex real-world environment characterized by a multi-room layout and cluttered objects. To evaluate the system's efficiency, we profiled the computational resource consumption of the onboard hardware in Fig.~\ref{fig:first-page-image}, confirming the lightweight nature of our framework. 

Qualitative results demonstrate that the UAV successfully comprehended the semantic information of various regions within the environment. By leveraging the structured semantic and geometric information, the system effectively executed hierarchical decision-making. This capability allowed the UAV to avoid redundant exploration of irrelevant areas, significantly improving search efficiency. A visualization of the real-world experiment is shown in Fig.~\ref{fig:first-page-image}.

\subsubsection{Extended Capabilities}
Beyond zero-shot exploration, the constructed scene graph serves as a persistent spatial memory for downstream tasks. First, it enables \textbf{Known Environment Navigation}: abstract commands like ``I want to attend a meeting'' are executed by directly retrieving target semantic regions (e.g., Meeting Room) and efficiently navigating to the target via path searching on Spatial Connectivity Graph without re-exploration. Second, it supports \textbf{Scene Reasoning}: by aggregating object semantics within topological bounds, the system answers high-level queries such as ``Describe current area,'' synthesizing detailed natural language descriptions of the spatial layout for intuitive human-robot interaction.

\section{CONCLUSION}
In this letter, we presented a unified spatio-semantic scene graph framework that enables UAVs to perform Zero-Shot Object Navigation in unknown environments. Our system incrementally constructs a hierarchical graph fusing spatial topology with open-vocabulary semantics directly on onboard edge computing platforms, to support subsequent LLM understanding and decision-making. This approach effectively addresses the conflict between the computational demands of semantic perception and the strict resource constraints of aerial robots. For future work, we aim to enhance the local planner by incorporating Vision-Language-Action models. This integration will allow for fine-grained, semantically aware local decision-making, further improving search efficiency and interaction capabilities in complex scenes.

\addtolength{\textheight}{-12cm}   

\bibliographystyle{IEEEtran}
\bibliography{References}
 

\end{document}